\documentclass[10pt,twocolumn,letterpaper]{article}

\usepackage{iccv}
\usepackage{times}
\usepackage{epsfig}
\usepackage{graphicx}
\usepackage{amsmath}
\usepackage{amssymb}

\usepackage{float}
\usepackage{algorithmic}
\usepackage{algorithm}
\usepackage{array}
\usepackage{subcaption}

\usepackage[square,sort,comma,numbers]{natbib}
\newcommand{\pluseq}{\mathrel{+}=}
\newcommand{\f}{\mathbf{f}}
\newcommand{\kk}{\mathbf{k}}

\usepackage[breaklinks=true,bookmarks=false]{hyperref}

\iccvfinalcopy % *** Uncomment this line for the final submission

\def\iccvPaperID{****} % *** Enter the ICCV Paper ID here

\begin{document}
	
	%%%%%%%%% TITLE
	\title{Filter Bank Regularization of Convolutional Neural Networks}

	\author{Seyed Mehdi Ayyoubzadeh\\
		McMaster University\\
		Hamilton, Canada\\
		{\tt\small ayyoubzs@mcmaster.ca}
		\and
		Xiaolin Wu\\
		McMaster University\\
		Hamilton, Canada\\
		{\tt\small xwu@mcmaster.ca}
	}
	
	\maketitle
	%\thispagestyle{empty}

	%%%%%%%%% ABSTRACT
	\begin{abstract}
		Regularization techniques are widely used to improve the generality, robustness, and efficiency of deep convolutional neural networks (DCNNs). In this paper, we propose a novel approach of regulating DCNN convolutional kernels by a structured filter bank. Comparing with the existing regularization methods, such as $\ell_1$ or $\ell_2$ minimization of DCNN kernel weights and the kernel orthogonality, which ignore sample correlations within a kernel, the use of filter bank in regularization of DCNNs can mold the DCNN kernels to common spatial structures and features (e.g., edges or textures of various orientations and frequencies) of natural images. On the other hand, unlike directly making DCNN kernels fixed filters, the filter bank regularization still allows the freedom of optimizing DCNN weights via deep learning. This new DCNN design strategy aims to combine the best of two worlds: the inclusion of structural image priors of traditional filter banks to improve the robustness and generality of DCNN solutions and the capability of modern deep learning to model complex non-linear functions hidden in training data. Experimental results on object recognition tasks show that the proposed regularization approach guides DCNNs to faster convergence and better generalization than existing regularization methods of weight decay and kernel orthogonality.
		
	\end{abstract}
	
	%%%%%%%%% BODY TEXT
	\section{Introduction}
	
	\subsection{Regularization}
	Deep convolutional neural networks (DCNNs) have rapidly matured as an effective tool for almost all computer vision tasks ~\cite{1707.07012, 1409.4842, 1704.04861, 1608.06993, 1512.03385, 1409.1556}, including object recognition, classification, segmentation, superresolution, etc.  Compared with traditional vision methods based on analytical models, DCNNs are able to learn far more complex, non-linear functions hidden in the training images.  However, DCNNs are also known for their high model redundancy and susceptibility to data overfitting.  When having a very large number of parameters, DCNNs have high Vapnic-Chervonenkis (VC) dimension.  If trained on limited amount of samples from the data generating distribution, DCNNs are less likely to choose the correct hypothesis from the large hypothesis space \cite{Caruana00overfittingin}.
	In other words, there should be a balance between information in the training examples and the complexity of the network ~\cite{NIPS1991_563}.  The simplest model that could perform the task and generalize well on the real world data is the best one. But choosing the simplest model is not an easy task; simply reducing the number of parameters in a network runs the risk of removing the true hypothesis from the hypothesis space.  To prevent overfitting and improve the generalization capability, a common strategy is to use a complex model but put some constraints on the model to make it overlook noise samples.  In this way
	reducing the model complexity is not achieved by reducing the number of free parameters in the network, but rather by controlling the variance of the model and its parameters.  This strategy is known as regularization ~\cite{1804.08042}.
	
	Regularization methods for DCNNs fall into two categories.
	The regularization methods of the first category are procedural.  For example, Ioffe and Szegedy proposed to perform batch normalization in each layer to reduce the internal covariate shift in the network and improve the generalization and performance of the network ~\cite{1502.03167}.
	
	In ~\cite{JMLR:v15:srivastava14a} Srivastava et al.\ used a dropout technique to stochastically regularize the network; they showed that the dropout works like an ensemble of simpler models.
	Khan and Shah proposed a so-called Bridgeout stochastic regularization technique ~\cite{1804.08042}, and they proved that their method is equivalent to the $L_q$ norm penalty on the weights for a generalized linear model, where norm $q$ is a learnt hyperparameter.
	
	All regularization methods of the first category are implicit and quite weak in the sense that they do not directly act on the CNN loss function, nor they require the convolutional kernels to have any spatial structures.
	
\begin{figure}
		\begin{center}
			\includegraphics[width=\linewidth]{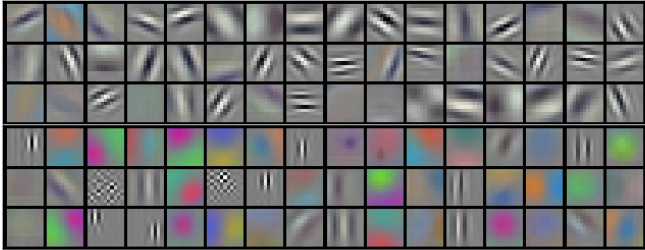}
			\caption{96 convolutional kernels of size $11 \times 11 \times 3$ learned by the first convolutional layer (AlexNet) \cite{Krizhevsky2017}}
			\label{fig:alexnetfilters}
		\end{center}
	\end{figure}

	The methods in the second category explicitly add a regularization term in the loss function to penalize the CNN weights.  One example is the weight decay method by A.\ Krogh and J.\ Hertz that penalizes large weights via the $\ell_2$ norm minimization ~\cite{NIPS1991_563}.  Among the published methods weight decay is the most common one to regularize CNNs.  For simple linear models, it can be shown, using Bayesian inference, that weight decay statistically means that the weights obey a multivariate normal distribution with diagonal covariance prior.  In this case, maximum a posterior probability (MAP) estimation with Gaussian prior on the weights is equivalent to the maximum likelihood estimation with the weight decay term \cite{Goodfellow-et-al-2016}.
	However, weight decay regularization can be justified only if the weights within a CNN kernel have no correlation with each other.  This assumption is obviously false, as it is well known that the CNN kernels, upon convergence, typically exhibit strong spatial structures.
	To illustrate our point, the kernels of the Alexnet after training are shown in Figure \ref{fig:alexnetfilters}, where the weights in a kernel are clearly correlated.
	
	Another form of penalty term in the cost function for regularizing CNN weights is the orthogonality of the kernels \cite{1703.01827}.  But the requirement of mutually orthogonal kernels also ignores the spatial structures.
	
	In summary, all existing CNN regularization methods overlook spatial structures of images.  This research sets out to rectify the above common problem.  Our solution to it is a novel approach of regularizing CNN convolutional kernels by a structured filter bank.  The idea is to
	encourage the CNN kernels to conform to common spatial structures and features (e.g., edges or textures of various orientations and frequencies) of natural images.  But this is different from simply making the CNN kernels fixed structured filters; the filter bank regularization still allows the CNN filters to be fine-tuned based on input data.  This new CNN design strategy aims to combine the best of two worlds: the inclusion of structural image priors of traditional filter banks to improve the robustness and generality of CNN solutions and the capability of modern deep learning to model complex non-linear functions hidden in training data.
	More specifically, our technical innovations are
	\begin{itemize}
		\item Considering a convolutional kernel as a set of correlated weights and penalize them based on their structural difference from adaptively chosen reference 2D filters.
		\item Using filter banks as guidance for the convolutional kernels of DCNNs but at the same time allowing the kernel weights to deviate from the reference filters, if so required by data.
	\end{itemize}
	
	The remainder of the paper is structured as follows.  The next section briefly review related works.  Section 3 is the main technical body of this paper, in which we present the details and justifications of the proposed new regularization method.  In Section 4, we report experimental results on object recognition tasks.  The proposed regularization approach is shown to lead to faster convergence and better generalization than existing regularization methods of weight decay and kernel orthogonality.
	
	\subsection{Related Work}
	
	In ~\cite{1703.01827} Xie et al.\ proposed an orthonormal regularizer for each layer of the CNNs, as a means to improve the accuracy and convergence of the network.  Except being free of redundancy, orthogonal kernels do not consider spatial correlation between the weights within a given kernel, and hence irrespective of spatial structures.
	Some attempts were made to reduce the complexity of model by including priors in the kernels.
	Bruna and Mallat proposed a method called convolutional scattering networks in which they used fixed cascaded wavelets to decompose images \cite{1203.1513}.  Although the method had good performance on specific datasets, it reduces the capability of CNNs.  The wavelet prior is too rigid to effectively characterize a great variety of unknown image structures.
	%could adversely affect the CNN especially when we have a relatively large dataset since in large datasets, there are lots of unknown variations which are not considered in hand crafted filters.
	Similarly, Chan et. al. in \cite{1404.3606} proposed a network architecture called PCANet in order to create filter banks in the layers based on a PCA decomposition of input images.  This method can learn convolutional kernels from the inputs, but the output cannot affect the filter bank design.  This is in conflict with the design objective of DCNNs, which is to learn convolutional kernels with respect to outputs not just inputs.  For instance, for classification tasks, the goal is to learn conditional probabilities of output data in relation to the input.
	
	%n fact reduces the capability of the CNN a lot. Because an important part of the CNN is to learning the kernels regarding the outputs not just inputs, this comes from the fact that we want learn conditional probabilities specifically in classification task.
	
	To gain flexibility over the scattering network and also to use the wavelet features, Jacobson et. al.\ proposed a method called structured receptive fields \cite{1605.02971}.  They make every kernel a weighted sum of filters in a fixed filter bank that consists of Gaussian derivative basis functions.  However, this method works under the assumption that every kernel filter is smooth and can be decomposed into a compact filter basis \cite{1605.02971}, which may not hold in all layers.
	%Although this linear combination reduces the variance of the model in practice also it could limit the CNN capacity, one of the basic assumption in this method is that every filter is smooth and can be decomposed into a compact filter basis \cite{1605.02971}, while this is a reasonable assumption for the kernels, but we could have non smooth kernels especially in the intermediate layers. Also, this method is not a regularization method and requires major change in the architecture of the CNN to apply this model. Moreover, we outperformed this method on CIFAR-10 \cite{CIFAR10} dataset.
	In \cite{1803.11405}, Keshari et al.\ published a method to learn a dictionary of filters in an unsupervised manner.  Then they made DCNN convolutional kernels linear combinations of the dictionary words with weights optimized by training data.
	Although this method reduces number of parameters of the network significantly, it limits the network performance, because the dictionary is not fine tuned by the training dataset. The implementation of this network is not an easy task because it needs
	a customized backpropagation for updating the weights.
	%For mitigating this problem they also offered iterative approach in which the dictionary and strength parameters update simultaneously. However training this network could be time consuming since it has a two step training iteratively.
	By combining pre-determined filters to form DCNN kernels, both of the mentioned methods severely limit the solution space of convolutional kernels when optimizing the DCNN for the given task.
	
	Sarwar et al.\ proposed a combination of Gabor filters and learnable filters when choosing convolutional kernels of DCNN \cite{1705.04748}.  For each layer they fixed some filters to be Gabor filters and allowed others to be trained.
	Luan et al.\ proposed a so-called Gabor convolutional network (GCN) \cite{1705.01450}.  The Gabor filters are used to modulate convolutional kernels of the DCNN.  The modulated filter kernels are optimized via back propagation.
	These two methods try to take advantage of the spatial structures of Gabor filters, but they are not used in regularization as we do in this paper.
	
	\section{Proposed Method}

	In this section, we explain our new filter bank regularization (FBR) technique for DCNNs in detail.  In the FBR method, we include in the DCNN objective function a penalty term that encourages the convolutional kernels of the DCNN to approach some member filters of a filter bank.
	In addition to controlling the model complexity of DCNNs to prevent overfitting and expedite convergence, the FBR strategy has a multitude of other advantages.  1.\ It is a way of incorporating into the DCNN design priors of spatial structures that are interpretable and effective; 2.\ The filter bank approach allows the DCNN kernels to be chosen from a large pool of candidate 2D filters suitable for a given computer vision application; 3.\ It is a general regularization mechanism that can be applied to any DCNN architecture without any modification.

	\subsection{Filter banks}

	Filter banks have proven their effectiveness for extracting useful features to facilitate many computer vision tasks.
	Being a set of different filters a filter bank can be used as bases (often overcomplete) to decompose images into meaningful construction elements.  Arguably the best known filter bank used in computer vision is Gabor filter bank, as the family of Gabor filters are noted for their power to characterize and discriminate texture and edges, thanks to their parameterization in orientation and frequency.  This is why the Gabor filter bank is a main construct used in the development of our FBR method.  In addition to their mathematical properties, Gabor filters can also, in view of many vision researchers, model simple cells in the visual cortex of mammalian brains. \cite{Marcelja:80}

	The generic formula of Gabor filter is:
	\begin{equation}\label{eqn:gabor}
	g(x,y;\lambda,\theta,\psi, \sigma, \gamma) = exp(-\frac{x'^2+\gamma^2y'^2}{2\sigma^2})exp(i(2\pi\frac{x'}{\lambda}+\psi))
	\end{equation}
	where:
	\begin{align*}
	x'&=xcos(\theta)+ysin(\theta)\\
	y'&=-xsin(\theta)+ycos(\theta)
	\end{align*}
	
	Typically, the real part of this filter is used for filtering images.
	The Gabor filter enables us to extract orientation-dependent frequency contents of the image \cite{1705.01450}. Transforming an image with Gabor filter bank decomposes it in a way that can enhance the separation capability of the machine learning model between different classes.
	Also, using the Gabor filter bank may be justified cognitively as some researchers showed that simple visual cortex cells of mammals could be modeled by Gabor filters \cite{Daugman1985}.

\begin{figure}
		\centering
		\includegraphics[width=\linewidth]{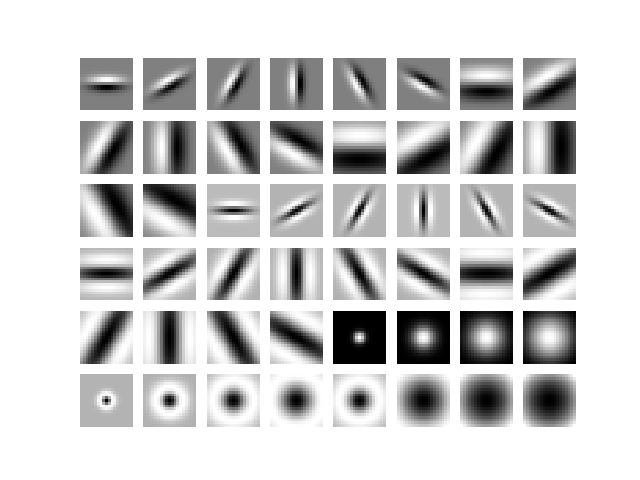}
		\caption{LM filter bank}
		\label{fig:filterbank_lm}
	\end{figure}

	To further enrich the DCNN model, we augment the Gabor filters in the regularization filter bank by adding the
	%try to combine traditional filter banks and DCNNs, we do this by proposing FBR.
	%In addition, there are some other useful filter banks such as
	Leung-Malik (LM) filter bank \cite{Leung2001} that has shown potential for extracting textures.  LM filter bank consists of first and second derivatives of Gaussians at 6 orientations and 3 scales, 8 Laplacian of Gaussian (LoG) filters, and 4 Gaussians filters. This filter bank is shown in Figure \ref{fig:filterbank_lm}.
	
	Using filter banks can increase robustness of the model for small varations. In the context of deep learning it has been shown that first layer kernels in the DCNNs that are trained on relatively large datasets such as VGGNet, ResNet and AlexNet are very similar to Gabor filters.
	Scale-space theory \cite{Witkin1987} gives a method for convolving an image with filters that have different scales, this method can be used to extract useful descriptors from general signals. Similarly, we try to use filter banks as a guidance for CNN filters, but as previously mentioned, the filter banks are suitable for general signals, so we guide DCNN kernels to be close to the filter bank. In what follows, we give detail about the implementation of this regularization.
	
	\subsection{Filter bank regularization as a Maximum A Posteriori Estimation (MAP) Problem}
	
	Using Bayesian statistics is a common approach to derive the regularized loss functions. We use MAP estimation to make the Bayesian posterior tractable. Consider the simple case of regression (${\rm I\!R}^n \rightarrow {\rm I\!R}$). The model parameter is $\boldsymbol{w}$ ($\boldsymbol{w} \in {\rm I\!R}^n$). The dataset contains $N$ pairs of datapoints denoted by $(\boldsymbol{x_1}, y_1), \cdots, (\boldsymbol{x_N}, y_N)$, in which  $\boldsymbol{x_i}$s are 1-D vectors and $y_i$s are scalers. In the presense of Gaussian noise we could write $y(\boldsymbol{x})=\hat{y}(\boldsymbol{x})+\boldsymbol{\epsilon}$, where  $\boldsymbol{\epsilon}$ is the Gaussian noise, and $\hat{y}$ is the model output. The conditional distribution of $y$ can be written as follows:
	\begin{align}
	P(y|\boldsymbol{x}) = \mathcal{N}(\hat{y}, I_N)
	\end{align}
	where $I_N$ is the $N \times N$ identity matrix. The MAP estimation for the model parameters $\boldsymbol{w}$ is defined by:
	
	\begin{align}
		\boldsymbol{w^*} & = \operatorname*{argmax}_{\boldsymbol{w}} P(\boldsymbol{w}|\boldsymbol{x},y) \nonumber \\ & = \operatorname*{argmax}_{\boldsymbol{w}}log(P(y|\boldsymbol{x}, \boldsymbol{w}))+log(P(\boldsymbol{w})) \end{align}
	
	where $P(\boldsymbol{w})$ is the prior distribution of the model parameters.
	Substituting (2) into (3) gives us:

    \begin{align}
    \boldsymbol{w^*} & = \operatorname*{argmax}_{\boldsymbol{w}} -(y-\hat{y})^T(y-\hat{y}) + log(P(\boldsymbol{w})) \nonumber \\
    & = \operatorname*{argmin}_{\boldsymbol{w}} \|y-\hat{y}\|_2^2  -log(P(\boldsymbol{w}))
    \end{align}
    
    Therefore, the cost function can be derived as follows:
	
	\begin{align}
	E(\boldsymbol{w}) = \|y-\hat{y}\|_2^2  -log(P(\boldsymbol{w}))
	\end{align}
 
 This result can be easily extended for 2-D datasets and parameters. As we discussed earlier in this paper, many researchers use the Gabor filters to model simple cells in the visual cortex of mammalian brains.  
  We Can use this information to presume a reasonable prior distribution for the model parameters. We assume that the model parameters have a Gaussian distribution around a vectorized filter $\boldsymbol{f}$ in the Gabor filter bank. In other words, we could write the  prior distribution of the parameters as follows:
     
    \begin{align}
    P(\boldsymbol{w}) = \mathcal{N}(\boldsymbol{f}, \frac{1}{\lambda} I_n)
    \end{align}
	
	$\lambda$ determines the deviation of the model kernel from $\boldsymbol{f}$. So, we can write the loss function as:
	
    \begin{align}
	E(\boldsymbol{w}) = \|y-\hat{y}\|_2^2  + \lambda \|\boldsymbol{w}-\boldsymbol{f}\|_2^2
	\end{align}
	
	\subsection{Kernel regularization using a filter bank}
	
	%Each kernel in DCNN is a three dimensional tensor. In each iteration of the learning process, we choose a reference filter for each channel in that tensor and then penalize that channel of the kernel based on the distance from the reference filter in the filter bank. The details are as follows.
	%Assume DCNN has $L$ convolutional layers. Let number of filters in $l$th layer be $M$, and the dimension of each filter in this layer is where . we denote the set of kernels in $l$th layer with .
	
	Denote by $\mathcal{F} = \{\f_1,\f_2,\cdots,\f_N\}$ a 2D filter bank of dimension $W \times H$.
	For a DCNN of $L$ convolution layers, let $M_l$ be the number of kernels in layer $l$, $1 \leq l \leq L$.  Denote the $M_l$ convolutional kernels of layer $l$ by $\kk_{l,1}, \kk_{l,2}, \cdots, \kk_{l,M_l}$.
	Each kernel in the DCNN is a three-dimensional tensor of dimension $W\times H \times D_l$, where $D_l$ is the number of channels in layer $l$.
	In each iteration of the learning process, for layer $l$ of the DCNN we find the filter in the filter bank $\mathcal F$ that best matches kernel $m$ in channel $d$, namely,
	\begin{equation}
	\label{eqn:matchfilter}
	\f^*_{l,m,d} = argmin_{\f \in \mathcal{F}} \|\f - \kk_{l,m,d}\|_2
	\end{equation}
	where $\kk_{l,m,d}$ is the 2D cross section of the 3D kernel $\kk_{l,m}$ in channel $d$.
	Accordingly, the FBR regularizer produces the penalty term
	\begin{equation}
	\label{eqn:regulperchannel}
	\Omega_{l,m,d}= \| \kk_{l,m,d} - \f^*_{l,m,d} \|_2^2
	\end{equation}
	Therefore, the total loss function of the DCNN is
	\begin{eqnarray}
	\label{eqn:finalloss}
	E(\mathbf{w})=\frac{1}{N}\sum_{n=1}^{N} L(x^{(n)}, y^{(n)},\mathbf{w}) +
	\\ \nonumber
	\lambda\sum_{l=1}^{L}
	\sum_{m=1}^{M_l}\sum_{d=1}^{D_l}\Omega_{l,m,d}
	\end{eqnarray}
	where $\mathbf{w}$ is the total weights of the DCNN, $L(x^{(n)}, y^{(n)},w)$ is the per sample classification loss for the input $x^{(n)}$ and corresponding output $y^{(n)}$.
	
	%$M_l$ is the number of kernels in $l$th layer of the DCNN and $D_{m,d}^{(l)}$ is the number of channels of those kernels. The $l$th layer of the DCNN is shown below:
	
\begin{figure}
		\centering
		\includegraphics[scale=0.3]{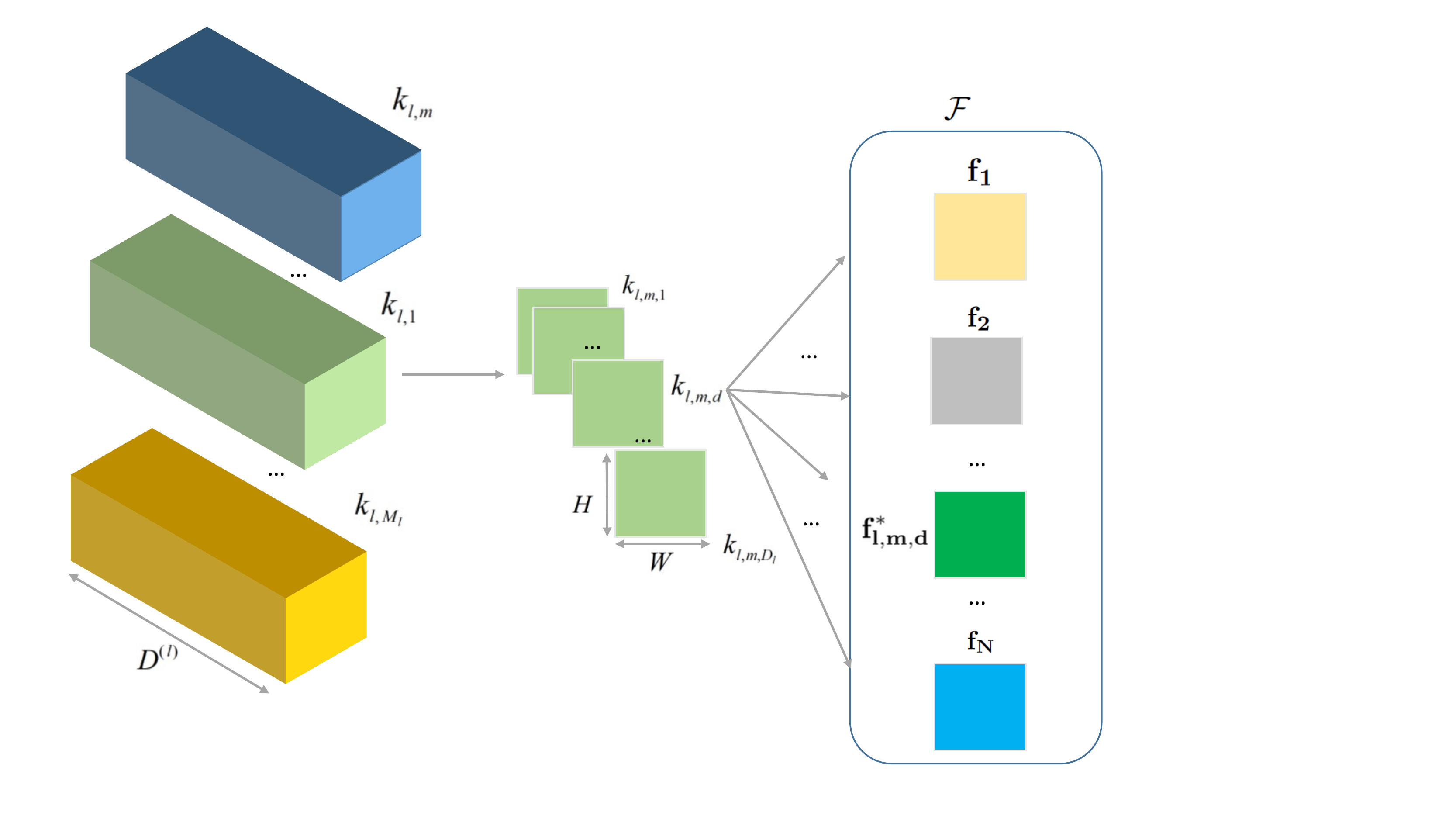}
		\caption{Finding best matches in $\mathcal{F}$ for kernels in layer $l$}
		\label{fig:interaction}
	\end{figure}

	The interactions between the DCNN convolutional kernels and the regularization filter bank are depicted in Figure \ref{fig:interaction}.
	
	As one could see, in FBR, kernels choose the reference regularizer adaptively, moreover, the reference filters are well structured in the spatial domain. The proposed algorithm is shown in Algorithm \ref{algo1}.
	
	\begin{algorithm}
		\caption{DCNN regularization using FBR}
		\begin{algorithmic}
			\label{algo1}
			\FOR{each iteration}
			\STATE $reg \leftarrow 0$
			\FOR{all layers in the DCNN}
			\STATE $l \leftarrow layer\;index$
			\FOR{all filters in $l$th layer}
			\STATE $m \leftarrow kernel\;index$
			\FOR{all channels in $m$th kernel}
			\STATE $d \leftarrow channel\;index$
			\STATE $i \leftarrow argmin_i{\|k[l][m][d] - f[i]\|_2}$
			\STATE $reg\pluseq\|k[l][m][d] - f[i]\|_2^2$
			\ENDFOR
			\ENDFOR
			\ENDFOR
			\STATE $(X,Y)$ $\leftarrow$ Select $N$ random samples from the dataset
			\STATE $L_N(X,Y,\mathbf{w})$ $\leftarrow$ Calculate average classification loss on $(X,Y)$
			\STATE $E(\mathbf{w})$ $\leftarrow$ Add $reg$ to $L_N(X,Y,\mathbf{w})$
			\STATE $\mathbf{w}$ $\leftarrow$ Update $\mathbf{w}$ via backpropagation
			\ENDFOR
		\end{algorithmic}
	\end{algorithm}

	\subsection{Adding orthogonality regularization}
	
	Due to the random initialization of the DCNN, it is likely that some of the kernels tend to select the same reference filter from the filter bank $\mathcal{F}$.
	This can create redundant or correlated kernels after DCNN training stage.  To resolve this issue, we introduce an orthogonality regularization term \cite{1709.06079} to encourage uncorrelated kernels.  Adding the orthogonality term can change the reference regularizer filters and as a result, enables the DCNN to learn a richer set of kernels.  Letting $\mathbf{w}_l$ be the kernel weight matrix of DCNN layer $l$ in which each column is a vectorized kernel, the orthogonality regularization term for this layer can be written as
	\begin{equation}
	\psi_l = \|\mathbf{w}_l^T\mathbf{w}_l-\mathbf{I}\|_F
	\end{equation}
	where $\mathbf{I}$ is the identity matrix and $F$ denotes Frobenius norm.
	By adding the orthgonality regularization, we can rewrite the final loss function as follows
	\begin{multline}
	E(\mathbf{w})=\frac{1}{N}\sum_{n=1}^{N}L(x^{(n)}, y^{(n)},\mathbf{w}) \\ + \lambda\sum_{l=1}^{L}
	\sum_{m=1}^{M_l}\sum_{d=1}^{D_l}\Omega_{l,m,d}
	+ \gamma \sum_{l=1}^{L}\psi_l
	\end{multline}
	where $\psi_l$ is the orthogonality regularization for layer $l$ of the DCNN.
	
	\begin{figure}
		\centering
		\includegraphics[width=\linewidth]{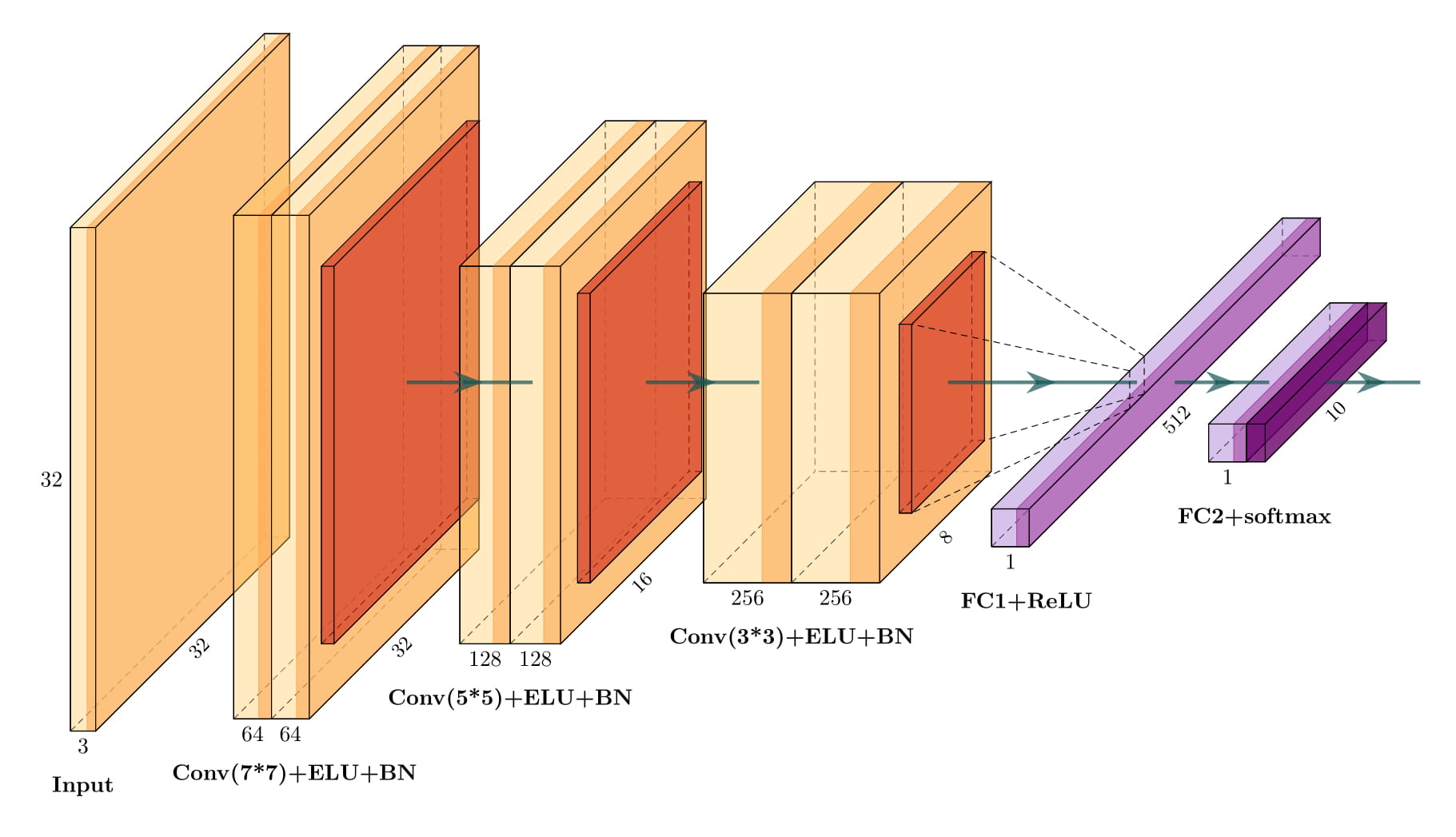}
		\caption{CNN baseline model}
		\label{fig:baselinecnn}
	\end{figure}
		
	\section{Experiments and Discussions}
	
\begin{figure}
		\centering
		\includegraphics[width=\linewidth]{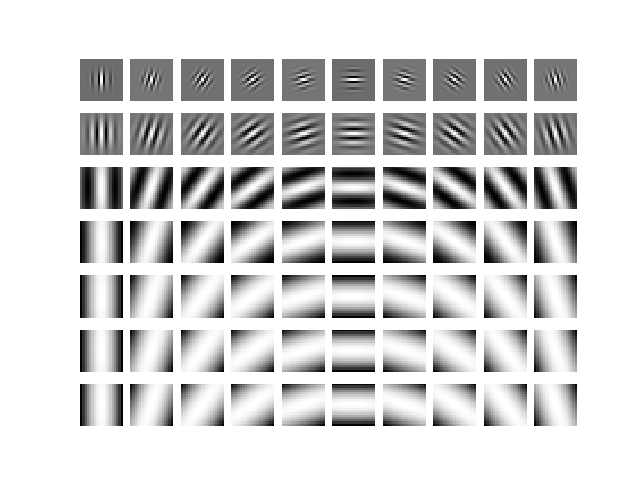}
		\caption{Gabor filter bank ($7$ orientations and $10$ frequencies)}
		\label{fig:filterbank_gabor}
	\end{figure}

	We implemented the proposed FBR method with classification DCNNs and evaluated its performances in comparison with existing regularization methods, including $\ell_1$, $\ell_2$ penalty norm on the weights and pure orthogonality regularization.
	Two commonly used benchmark datasets CIFAR 10 \cite{CIFAR10} and Caltech-101
	\cite{caltech101} are used in our evaluations.
	In our experiment setup, the filter bank is the union of the Gabor and LM filter banks.  These filter banks are shown in Figure \ref{fig:filterbank_lm} and \ref{fig:filterbank_gabor}.  We designed the Gabor filter bank using 10 different orientations and 7 frequencies with $\sigma=\frac{1}{f}$ resulting 70 Gabor member filters.
	
	\subsection{Results on MNIST benchmark}
	First, we report our experimental results on the benchmark dataset MNIST \cite{lecun-mnisthandwrittendigit-2010}. We apply Double Soft Orthogonality (DSO) regularization \cite{1810.09102} to the DCNN model. The model consists of $5$ convolution layers. The first $3$ convolutional layers are regularized. The learning rate of $0.001$ is used, and we make it half every $10$ epoch. We compare our method with Gabor Convolutional Networks (GCNs). As one could see in table \ref{table:mnist_results}, FBR has the best accuracy with much fewer parameters in comparison with GCNs.
	
		\begin{table}
		\begin{center}
			\footnotesize
			\begin{tabular}{ | c | c | c |}
				\hline
				\footnotesize Reg. Type & \footnotesize Err. (\%) & \footnotesize \#Params (M) \\ \hline
				\centering \small Baseline & \centering $0.52$ & $0.61$ \\ \hline
				\centering \small Ortho. ($\gamma=10^{-5}$)  & $0.49$ & $0.61$ \\ \hline
				\centering \small Ortho. ($\gamma=0.01$) & $0.53$ & $0.61$ \\ \hline
				\centering \small GCN4 (with $3 \times 3$)   & $0.56$ & $0.78$ \\ \hline
				\centering \small GCN4 (with $5 \times 5$)  & $0.48$ & $1.86$ \\ \hline
				\centering \small GCN4 (with $7 \times 7$)  & $0.42$ & $3.17$ \\ \hline
				\centering \small FBR ($\lambda=0.0001, \gamma=0.0001$)  & $0.40$ & $0.61$ \\ \hline
				\centering \small \textbf{FBR ($\mathbf{\lambda=0.0001}, \mathbf{\gamma=0.0}$)}    & $\mathbf{0.34}$ & $\mathbf{0.61}$ \\ \hline
			\end{tabular}
		\end{center}
		\caption{\small Error percentage on test dataset (MNIST)}
		\label{table:mnist_results}
	\end{table}

	\subsection{Performance evaluation on CIFAR-10 benchmark}
	
	%We compared the FBR method with commonly used regularization techniques such as $\ell_1$, $\ell_2$ penalty norm on the weights and pure orthogonality regularization.
	
	we report our experimental results on the benchmark dataset CIFAR-10.
	CIFAR-10 contains $50000$ training images and $10000$ testing images from $10$ different categories. The images dimensions are $32 \times 32$.
	The architecture that we used for DCNN is shown in Figure \ref{fig:baselinecnn}. We applied regularization on the $7 \times 7$ and $5 \times 5$ convolution kernels and trained the model for $300$ epochs, using the RMSProp optimizer with learning rate $10^{-3}$ and decay of $10^{-6}$. The batch size was set to $128$, and data augmentation was used.  We also used step decay to half the learning rate after every $25$ epochs.  It is worth mentioning that we employed regularization only for $4$ layers of the DCNN when dimension of kernels were larger than $3 \times 3$, in order to have an effective filter bank with reasonable representational capability. To make the comparisons fair, we used the same weight initialization for all experiments.
	
	\subsubsection{Discussions}
	
\begin{figure}
		\begin{center}
			\includegraphics[width=\linewidth]{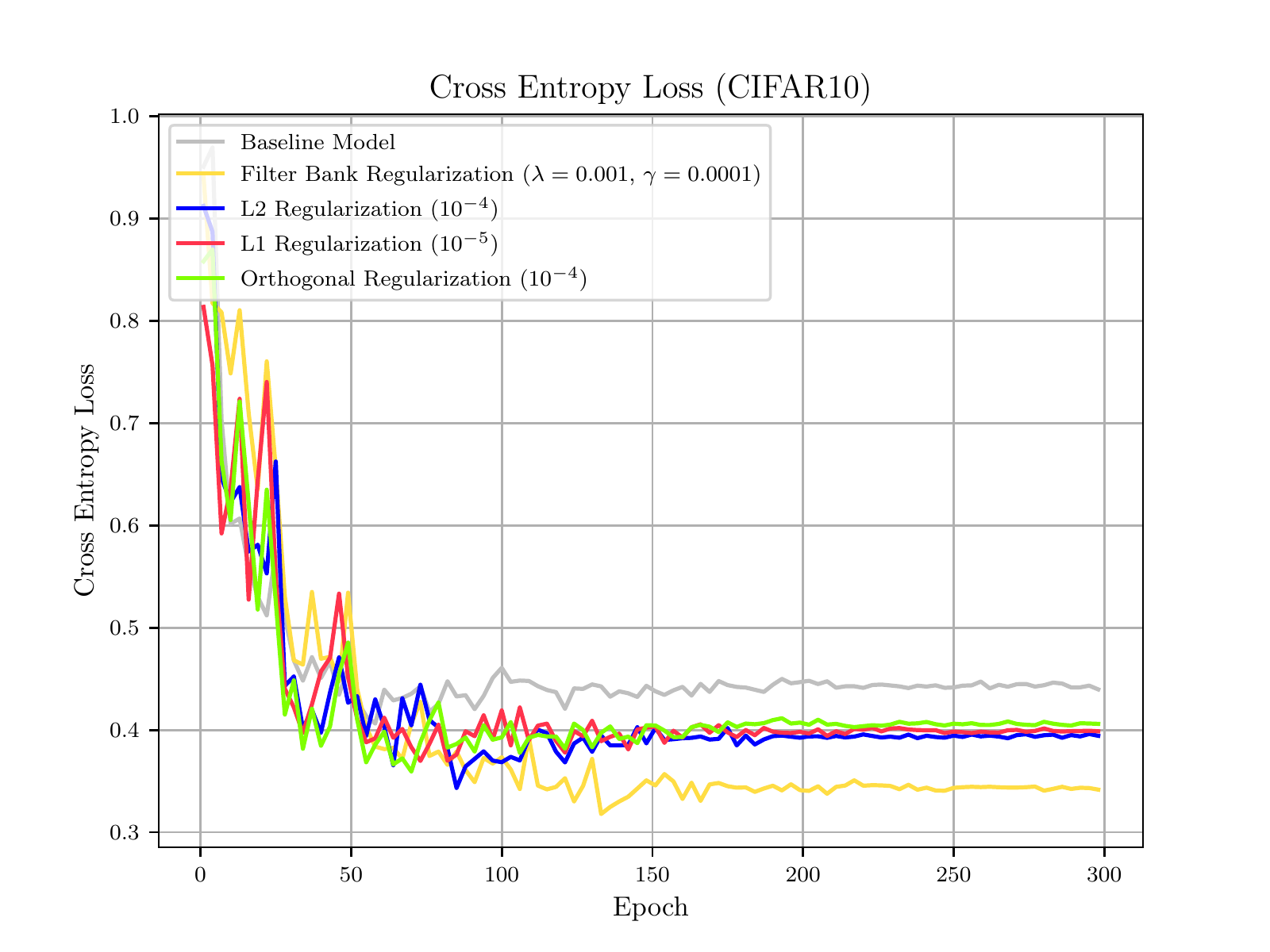}
			\caption{Cross entropy (CE) loss on test dataset (CIFAR-10)}
			\label{fig:testlosscifar}
		\end{center}
	\end{figure}

	The cross entropy loss results of different regularization methods on the test dataset are plotted in Figure \ref{fig:testlosscifar}. In the figure, the baseline curve is for the classification DCNN of Figure 4 without any regularization.  As shown, the FBR method has the lowest cross entropy loss ($0.341$) among all tested methods.  An interesting observation is that, the reduction in cross entropy with respect to the baseline is almost the same for the orthogonal regularization, $\ell_1$ and $\ell_2$ regularization.  And the FBR method can reduce the cross entropy further from the above three methods by approximately same margin.
	Additionally in Table \ref{table:cifar10_results}, we tabulate the experiments results in more details, including both the classification accuracy and cross entropy numbers in relationship to hyperparameters $\lambda$ and $\gamma$.  The table demonstrates that the FBR method outperforms all other regularization methods, in both the classification accuracy and cross entropy loss for suitable $\lambda$ and $\gamma$.
	
\begin{figure}
		\centering
		\begin{subfigure}[b]{0.33\linewidth}
			\includegraphics[width=\linewidth]{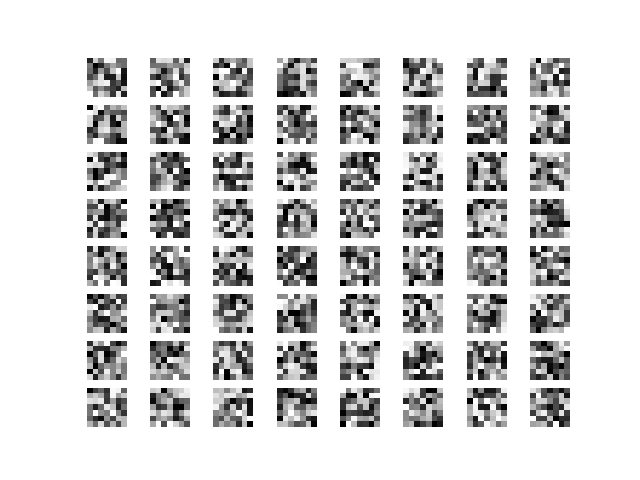}
			\caption{Initialzed kernel weights.}
		\end{subfigure}
		\begin{subfigure}[b]{0.33\linewidth}
			\includegraphics[width=\linewidth]{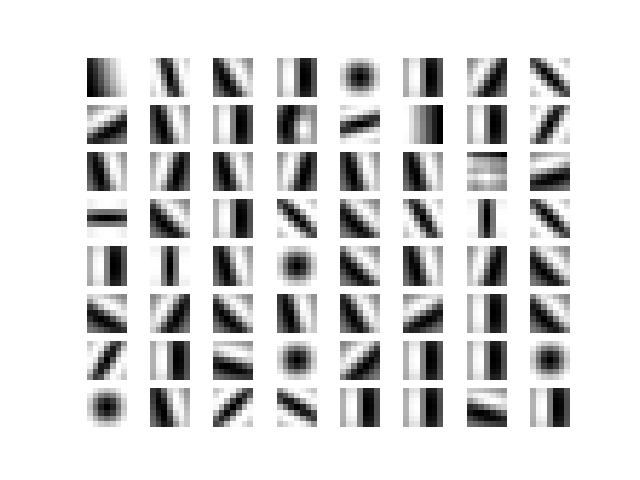}
			\caption{Learned kernel weights ($\gamma=0$).}
		\end{subfigure}
		\begin{subfigure}[b]{0.33\linewidth}
			\includegraphics[width=\linewidth]{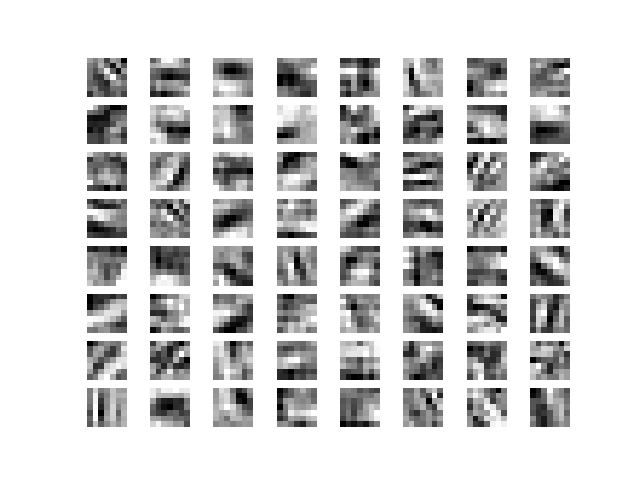}
			\caption{Learned kernel weights ($\gamma = 10^{-3}$)}
		\end{subfigure}
		\caption{Effects of $\gamma$ on the learned convolutional kernel weights (the second layer of the DCNN.)}
		\label{fig:filters_shape}
	\end{figure}

	Also, one can see the effects of $\gamma$ on the spatial structures of DCNN kernels in Figure \ref{fig:filters_shape}.  Emphasizing the orthogonality of the DCNN kernels can reduce the degree of kernel redundancy, i.e., preventing similar kernels from being chosen.
	
	\subsection{Effect of kernel size}
	
	In the FBR method, as opposed to other regularization techniques, the DCNN kernel size affects both the DCNN architecture and the regularization filter bank $\cal F$.  Increasing the kernel size improves the spatial resolution of the filter bank while sacrificing the locality of the feature maps.  In practice, we need to trade off between the spatial resolution and the locality by varying the kernel size.  Decreasing the kernel size can reduce the representational power of the filter bank, but on the other hand, improve the locality.
	To examine how much the kernel size can affect the DCNN model, we trained the DCNN baseline with different kernel sizes.
	The results of these experiments are shown in Table \ref{tab:Ksizeeffect}.
	
	%Kernel size is definitely an important factor, because as opposed to common regularization techniques, in FBR a reference filterbank is used for regularization. To examine how much kernel size can affect the DCNN model, we trained the DCNN baseline with different kernel sizes. Intuitively, there is trade off in changing the kernel size. Decreasing the kernel size can reduce the representational power of the filter bank, on the other hand, increase the locality.
	
	\begin{table}
		\begin{center}
			\begin{tabular}{ |  c | c | c|}
				\hline
				\small Kernel Size & \small Accuracy (\%) & \small CE Loss \\ \hline
				$5$ & $90.39$ & $0.350$ \\ \hline
				$\mathbf{7}$ & $\mathbf{90.90}$ & $\mathbf{0.346}$ \\ \hline
				$9$ & $90.24$ & $0.384$ \\ \hline
				$11$ & $88.41$ & $0.426$ \\ \hline
				
			\end{tabular} 	
		\end{center}
		
		\caption{\label{tab:Ksizeeffect} The effect of the kernel size in the first DCNN layer on the performance.}
		
	\end{table}
	
	As one can see, the aforementioned trade-off creates an optimal kernel size for the DCNN. In fact, this can be achieved by optimizing a hyperparameter in a cross validation approach.
	
	\begin{table}
		\begin{center}
			\footnotesize
			\begin{tabular}{ | c | c | c | c | c| c |}
				\hline
				\footnotesize Reg. Type & $\gamma$ & $\lambda$ & \small $L_1$ or $L_2$ \tiny Coeff. & \footnotesize Acc. (\%) & \footnotesize CE Loss \\ \hline
				\centering \small Baseline & \centering  $0$ & \centering $0$ & \centering $0$ & \centering $89.43$ & $0.439$ \\ \hline
				\centering \small $L_1$  & $0$ & $0$ & $10^{-6}$ & $89.93$ & $0.417$ \\ \hline
				\centering \small $L_1$  & $0$ & $0$ & $10^{-5}$ & $89.89$ & $0.398$ \\ \hline
				\centering \small $L_1$  & $0$ & $0$ & $10^{-4}$ & $87.87$ & $0.432$ \\ \hline
				\centering \small $L_2$  & \centering $0$ & $0$ & $10^{-5}$ & $89.32$ & $0.465$ \\ \hline
				\centering \small $L_2$  & $0$ & $0$ & $10^{-4}$ & $90.75$ & $0.394$ \\ \hline
				\centering \small $L_2$  & $0$ & $0$ & $10^{-3}$ & $90.51$ & $0.347$ \\ \hline
				\centering \small Ortho. & $10^{-3}$ & $0$ & $0$ & $90.55$ & $0.393$ \\ \hline
				\centering \small Ortho. & $10^{-4}$ & $0$ & $0$ & $90.23$ & $0.405$ \\ \hline
				
				\centering \small \textbf{FBR} & $10^{-2}$ & $10^{-5}$ & $0$ & $\mathbf{91.06}$ & $0.358$ \\ \hline
				\centering  \small FBR & $10^{-4}$ & $10^{-5}$ & $0$ & $89.68$ & $0.472$ \\ \hline
				\centering \small FBR & $0$ & $10^{-5}$ & $0$ & $89.14$ & $0.435$ \\ \hline
				\centering \small FBR & $10^{-2}$ & $10^{-4}$ & $0$ & $90.89$ & $0.374$ \\ \hline
				\centering \small \textbf{FBR} & $10^{-4}$ & $10^{-3}$ & $0$ & $90.7$ & $\mathbf{0.341}$ \\ \hline
			\end{tabular}
		\end{center}
		\caption{\small Accuracy and cross entropy loss on test dataset (CIFAR-10)}
		\label{table:cifar10_results}
	\end{table}
	
	\subsection{Results on Caltech-101}
	
	As discussed above, applying a large kernel to very small images like CIFAR-10 ($32 \times 32$), can lead to poor locality.  To avoid the problem and evaluate the performance of the FBR method with larger kernel sizes, we conduct the above experiments using the 
	Caltech-101 dataset \cite{caltech101} and compare different regularization methods. 
	Caltech-101 has 101 categories and each class contains $40$ to $800$ images. We resized all of the images to $128 \times 128$ and used the DCNN baseline architecture with two extra max pooling at the first and third convolutional layers to control the number of DCNN parameters.
	
	\begin{table}
		\begin{center}
			\footnotesize
			\begin{tabular}{ | c | c | c | c | c| c |}
				\hline
				\footnotesize Reg. Type & $\gamma$ & $\lambda$ &  \small $L_1$, \small $L_2$ \tiny Coeff. & \footnotesize Acc. (\%) & \footnotesize CE Loss \\ \hline
				\centering \small Baseline & \centering  $0$ & \centering $0$ & \centering $0$ & \centering $72.35$ & $1.578$ \\ \hline
				\centering \small $L_1$  & $0$ & $0$ & $10^{-6}$ & $74.27$ & $1.561$ \\ \hline
				\centering \small $L_1$  & $0$ & $0$ & $10^{-5}$ & $70.81$ & $1.660$ \\ \hline
				\centering \small $L_1$  & $0$ & $0$ & $10^{-4}$ & $70.62$ & $1.505$ \\ \hline
				\centering \small $L_1$  & $0$ & $0$ & $10^{-3}$ & $56.45$ & $1.951$ \\ \hline
				\centering \small $L_2$  & \centering $0$ & $0$ & $10^{-6}$ & $71.69$ & $1.670$ \\ \hline
				\centering \small $L_2$  & $0$ & $0$ & $10^{-5}$ & $73.11$ & $1.659$ \\ \hline
				\centering \small $L_2$  & $0$ & $0$ & $10^{-4}$ & $72.84$ & $1.597$ \\ \hline
				\centering \small $L_2$  & $0$ & $0$ & $10^{-3}$ & $71.62$ & $1.469$ \\ \hline
				\centering \small Ortho. & $10^{-4}$ & $0$ & $0$ & $73.54$ & $1.567$ \\ \hline
				\centering \small Ortho. & $10^{-3}$ & $0$ & $0$ & $73.65$ & $1.654$ \\ \hline
				\centering \small Ortho. & $10^{-2}$ & $0$ & $0$ & $75.65$ & $1.453$ \\ \hline
				\centering \small Ortho. & $10^{-1}$ & $0$ & $0$ & $75.34$ & $1.437$ \\ \hline
				\centering  \small FBR & $10^{-1}$ & $10^{-5}$ & $0$ & $75.84$ & $1.448$ \\ \hline
				\centering \small FBR & $10^{-2}$ & $10^{-5}$ & $0$ & $74.15$ & $1.619$ \\ \hline
				\centering \small \textbf{FBR} & $10^{-3}$ & $10^{-4}$ & $0$ & $\mathbf{76.65}$ & $1.556$ \\ \hline
				\centering \small \textbf{FBR} & $10^{-3}$ & $5*10^{-5}$ & $0$ & $75.72$ & $\mathbf{1.410}$ \\ \hline
				\centering \small FBR & $10^{-2}$ & $10^{-4}$ & $0$ & $75.84$ & $1.480$ \\ \hline
			\end{tabular}
		\end{center}
		\caption{\small Accuracy and cross entropy loss on test dataset (Caltech-101)}
		\label{table:caltech101_results}
	\end{table}

	The experimental results with the Caltech-101 dataset are presented in Table \ref{table:caltech101_results}.  By comparing Table \ref{table:caltech101_results} with Table 2 (CIFAR-10), we can see that not only the FBR method achieves the best performance in the comparison group, but also its performance gain over others increases by a significant margin with larger kernel sizes and higher resolution images.
	In other words, the FBR method is more advantageous on high resolution images of greater variations, because it can adapt the kernel size.  
	
\begin{figure}
		\includegraphics[width=\linewidth]{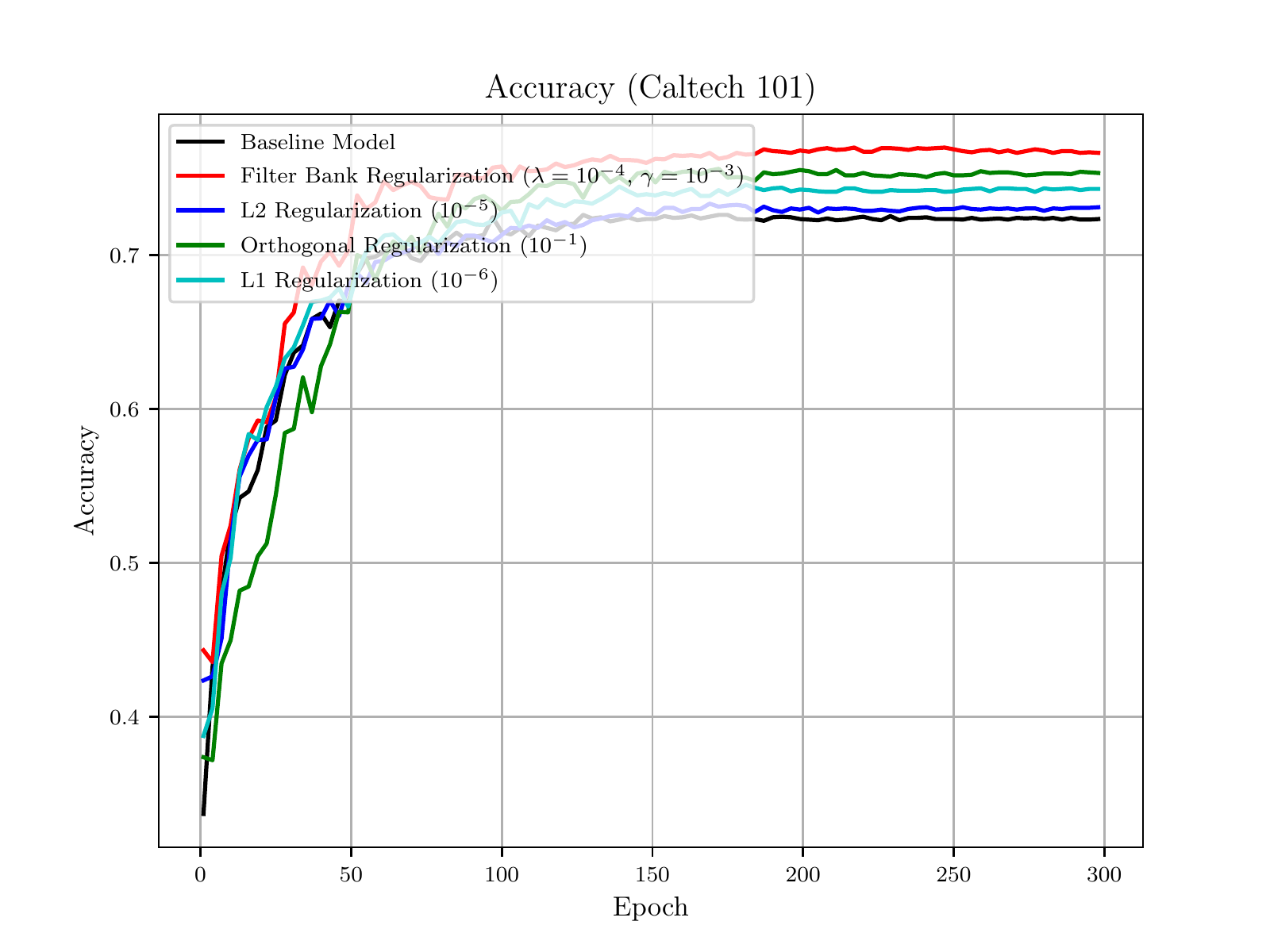}
		\caption{Accuracy on test dataset (Caltech-101)}
		\label{fig:testacccaltech}
	\end{figure}

	The classification accuracy results on the test dataset for different methods are plotted in Figure \ref{fig:testacccaltech}.  
	%The FBR method excelled other regularization tehcniques in both the classification accuracy and cross entropy loss for suitable $\lambda$ and $\gamma$.
	As one can observe in Figure \ref{fig:testacccaltech}, the $\ell_2$ regularization method improves the generalization of the model over the baseline by a very small amount, the $\ell_1$ regularization performs much better than the $\ell_2$ regularization.  The orthogonal regularization outperforms both $\ell_1$ and $\ell_2$, because the orthogonality prevents the choice of highly correlated kernels and promotes more diverse kernels to extract more novel features.
	
	% exploited the DCNN capability in a more effective way.
	
	\subsection{Large Size Image Classification}
	
\begin{figure}
		\includegraphics[width=\linewidth]{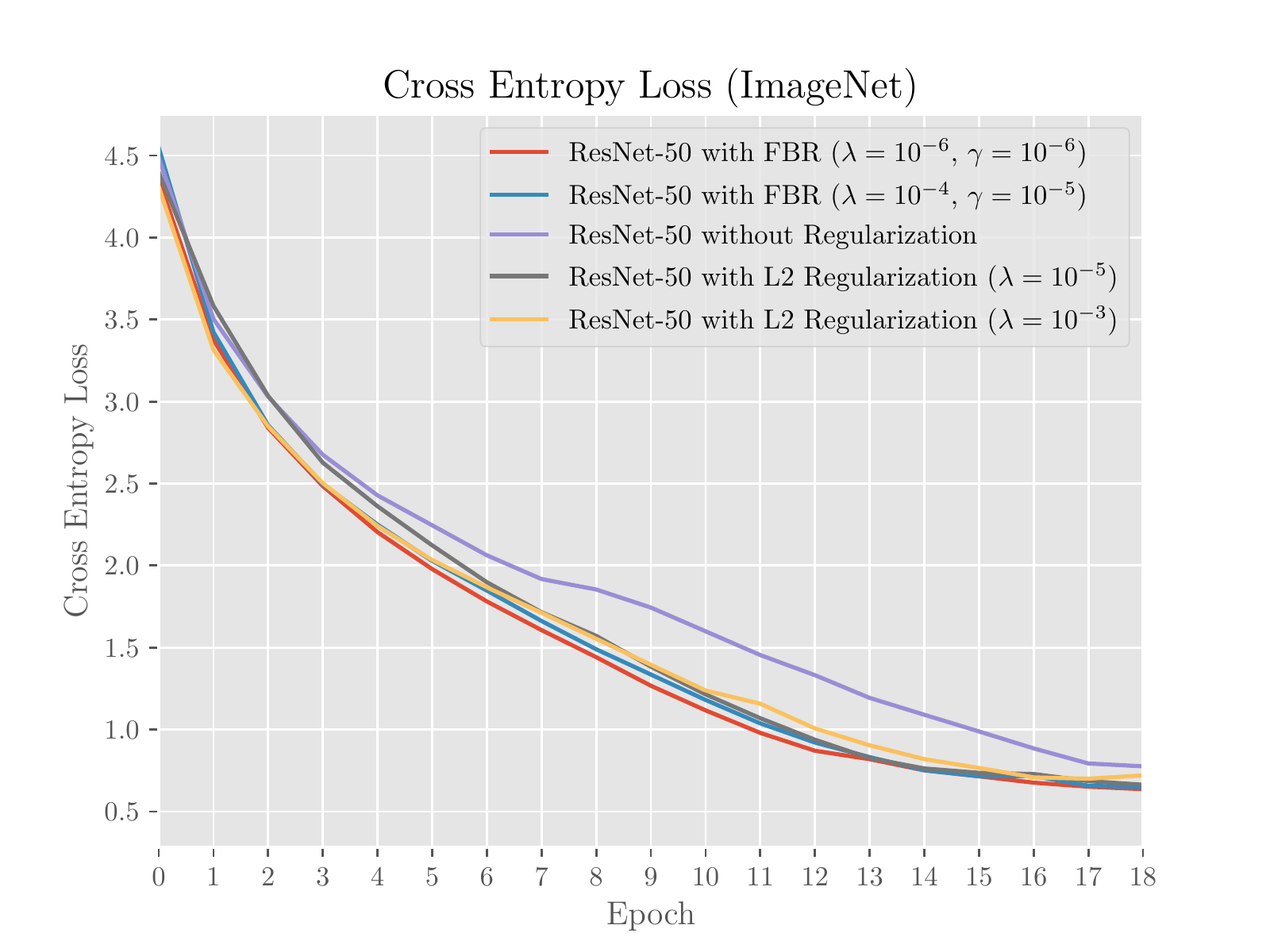}
		\caption{CE Loss on training dataset (ImageNet)}
		\label{fig:trainingimagenet}
	\end{figure}

\begin{figure}
		\includegraphics[width=\linewidth]{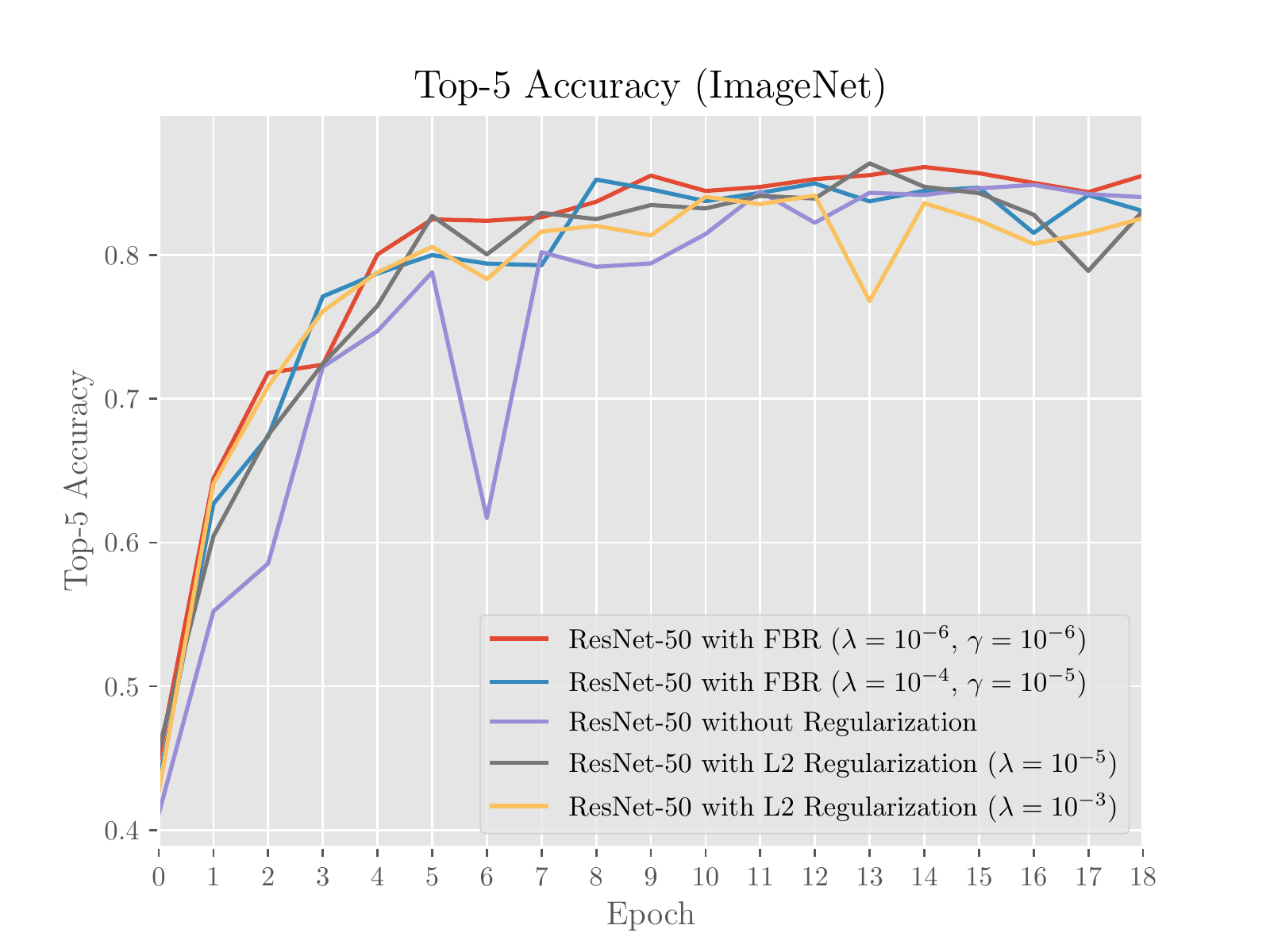}
		\caption{Top-5 Accuracy on validation dataset (ImageNet)}
		\label{fig:validationImageNet}
	\end{figure}

    To show that FBR is an effective method to regularize the DCNN on the large scale images as well as the small scale images, we use ImageNet \cite{ILSVRC15} dataset. It contains color images ($224 \times 224$) from $1000$ different objects. We use $100$ classes from the objects to train the DCNN. We use ResNet-50 architecture \cite{1512.03385} as our baseline model. We could see the training loss and top-5 accuracy on the validation data in the Figure \ref{fig:trainingimagenet} and \ref{fig:validationImageNet} respectively.

	As shown in Figure \ref{fig:trainingimagenet}, both models that regularized with FBR have the lower training loss in comparison with $L2$ regularization or baseline model without regularization. For the validation, as one could see in Figure \ref{fig:validationImageNet}, the model with FBR regularization has the best accuracy. In addition, the abrupt changes in validation accuracy for FBR model is less than other models.

	\subsection{Results of using a VGG-derived filter bank as the regularizer}
	
\begin{figure}
		\centering
		\includegraphics[scale=0.5]{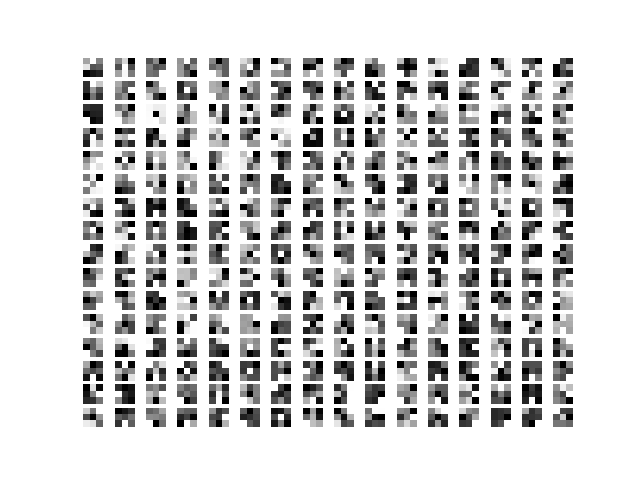}
		\caption{256 sampled filters ($3 \times 3$) from pretrained VGG16}
		\label{fig:vggfilterbank}
	\end{figure}

	We can also construct the regularization filter bank using the lower
	layer convolutional kernels of some pretrained DCNNs, for example,
	those of VGG16.
	As mentioned previously, Gabor filters do not work effectively if the filter kernel is small.  One way of creating a regularization filter bank of a small kernel size is to choose a subset of pre-trained VGG convolutional kernels at first few front layers.  Specifically, we randomly select 256 VGG16 kernels of the first two layers pre-trained on Imagenet to form the regularization filter bank, which is shown in Figure \ref{fig:vggfilterbank}.

	For comparison purposes, we train the DCNN of Figure \ref{fig:baselinecnn} to classify the Caltech 101 dataset, with the above VGG-derived filter bank regularization, the $\ell_2$ regularization, and without any weight regularization at all (the baseline), and compare the performances of these methods.
	
\begin{figure}
		\includegraphics[width=\linewidth]{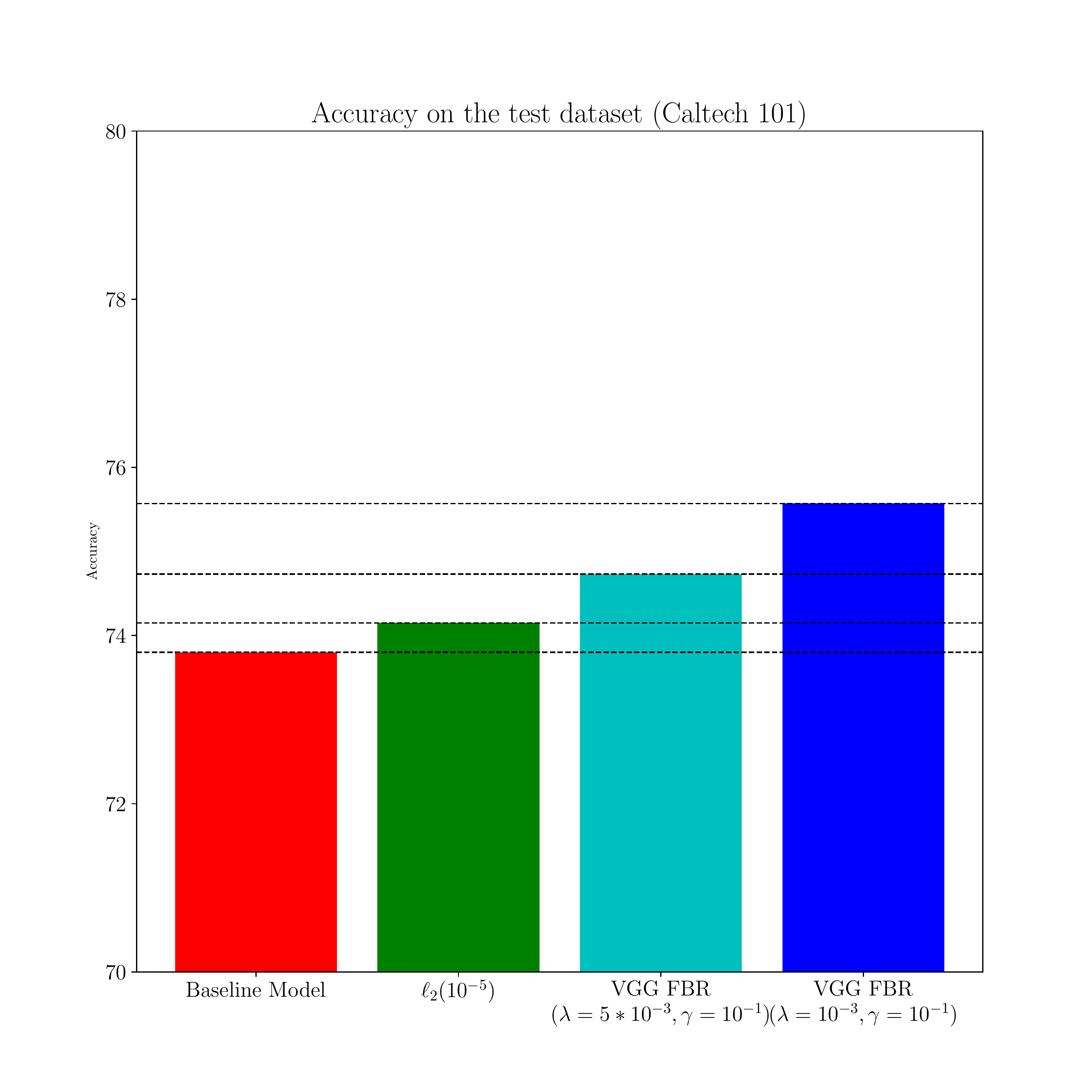}
		\caption{Accuracy on the test dataset (Caltech 101)}
		\label{fig:accc}
	\end{figure}

\begin{figure}
		\includegraphics[width=\linewidth]{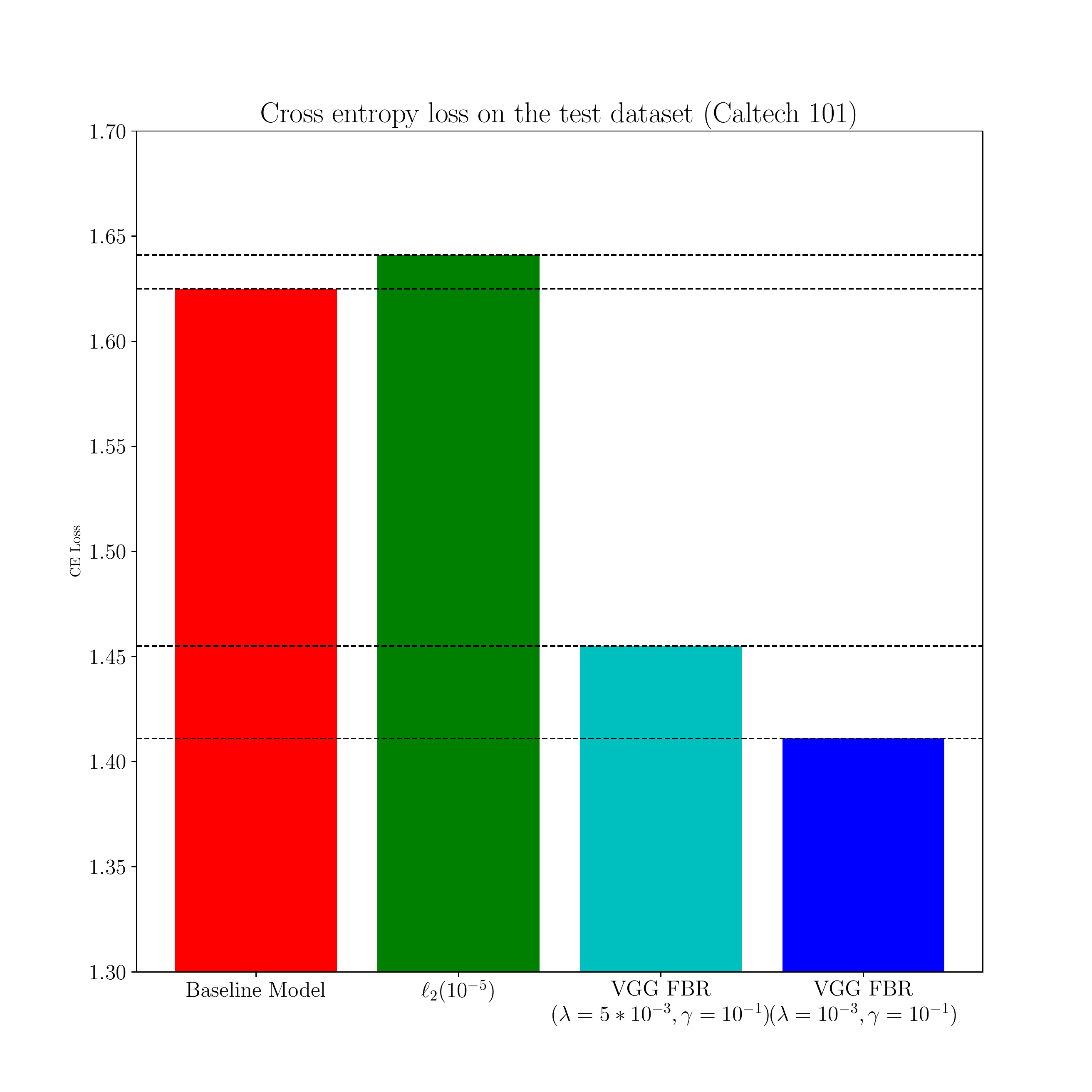}
		\caption{CE loss on the test dataset (Caltech 101)}
		\label{fig:lossc}
	\end{figure}

	The classification accuracy and cross entropy results are displayed in Figures \ref{fig:accc} and \ref{fig:lossc}, respectively.
	As shown, the DCNNs regularized by the VGG16-derived $3 \times 3$ filter bank outperform the  
	$\ell_2$-regularized DCNN and the baseline model without regularization.
	
	\subsubsection{Comparison of different regularization methods based on feature maps}
	
	In order to understand how the proposed FBR method works and its advantages over the other methods, we examine the DCNN feature maps generated under different and without regularizations. 
	
\begin{figure}
	\centering
	\begin{subfigure}[b]{0.45\linewidth}
		\includegraphics[width=\linewidth]{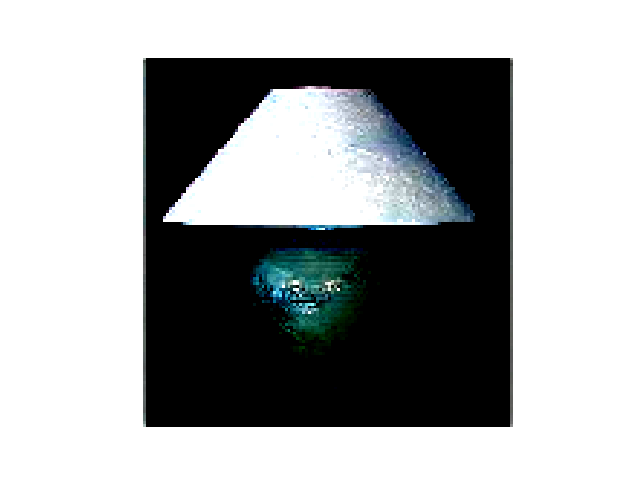}
		\caption{\scriptsize FBR prediction: "lamp", Baseline Prediction: "nautilus"}
	\end{subfigure}
	\begin{subfigure}[b]{0.45\linewidth}
		\includegraphics[width=\linewidth]{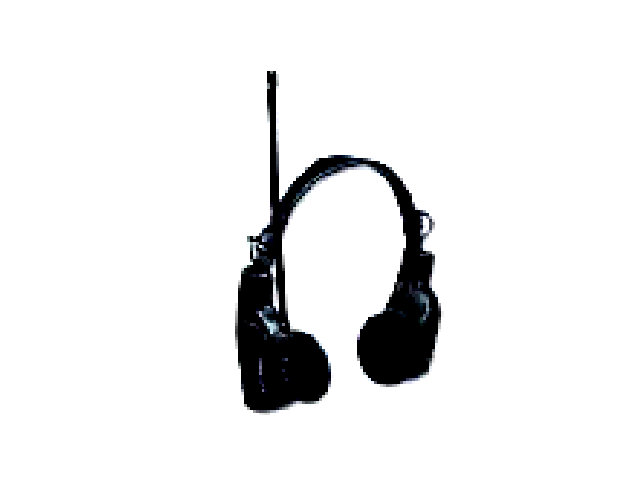}
		\caption{\scriptsize FBR prediction: "headphone", Baseline Prediction: "scissors"}
	\end{subfigure}
	\caption{Example images}
	\label{fig:testsample12}
	\end{figure}

	First we compare the baseline model without regularization and the FBR regularization method.
	Let us examine two examples from the Caltech 101 test dataset on which the baseline model misclassifies whereas the FBR method correctly classifies.  The two test images after normalization (mean subtracted and then divided by standard deviation) are shown in Figures \ref{fig:testsample12}.

\begin{figure}
		\centering
		\begin{subfigure}[b]{0.48\linewidth}
			\includegraphics[width=\linewidth]{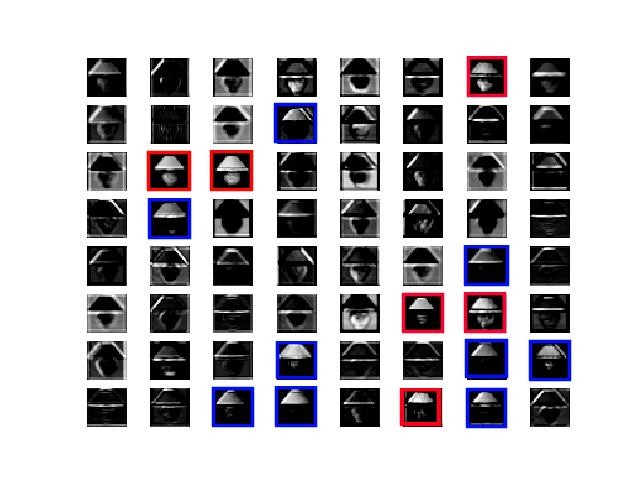}
			\caption{\scriptsize First layer, baseline model}
		\end{subfigure}
		\begin{subfigure}[b]{0.48\linewidth}
			\includegraphics[width=\linewidth]{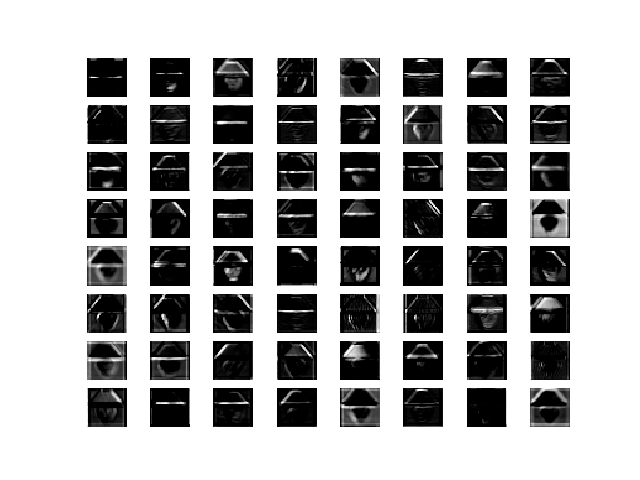}
			\caption{\scriptsize First layer, CNN using FBR}
		\end{subfigure}
		\begin{subfigure}[b]{0.48\linewidth}
			\includegraphics[width=\linewidth]{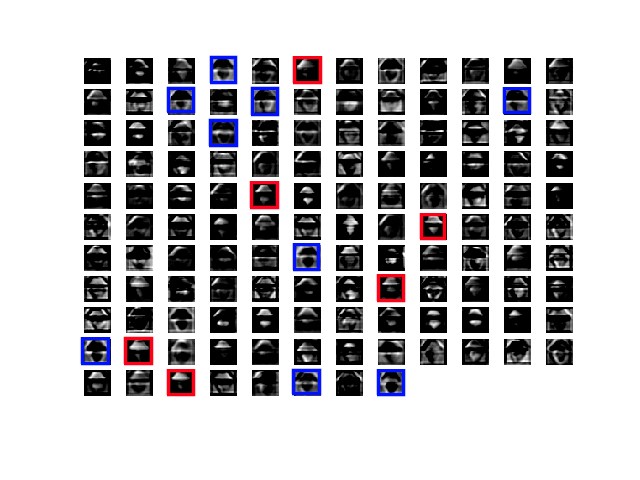}
			\caption{\scriptsize Third layer, baseline model}
		\end{subfigure}
		\begin{subfigure}[b]{0.48\linewidth}
			\includegraphics[width=\linewidth]{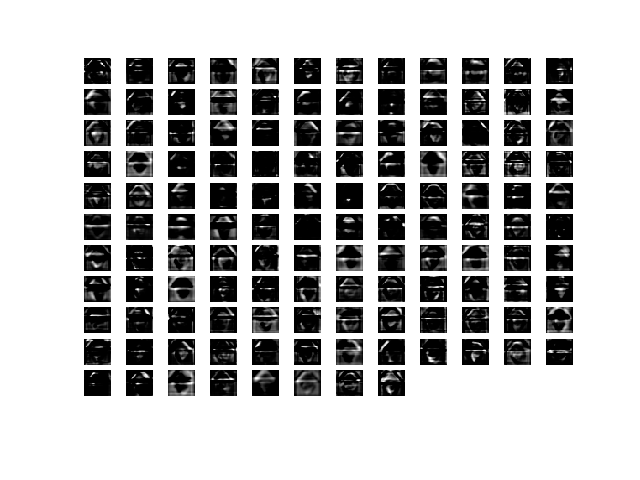}
			\caption{\scriptsize Third layer, CNN using FBR}
		\end{subfigure}
		\caption{Feature maps for the first test image (Very similar feature maps are marked with red and blue colors)}
		\label{fig:testsample1_featuremaps}
	\end{figure}

\begin{figure}
	\centering
	\begin{subfigure}[b]{0.48\linewidth}
		\includegraphics[width=\linewidth]{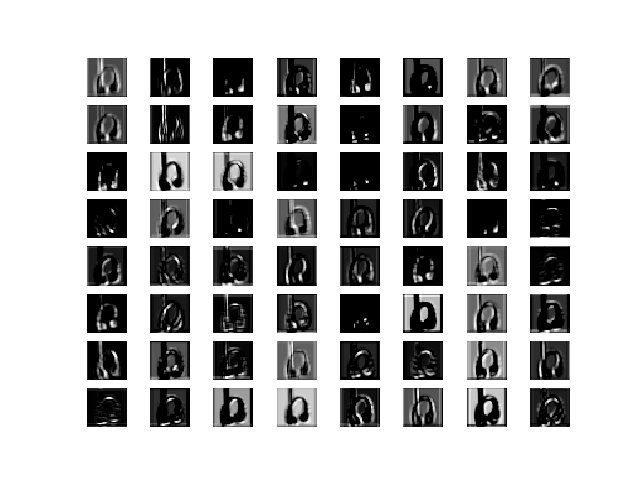}
		\caption{\scriptsize First layer, baseline model}
	\end{subfigure}
	\begin{subfigure}[b]{0.48\linewidth}
		\includegraphics[width=\linewidth]{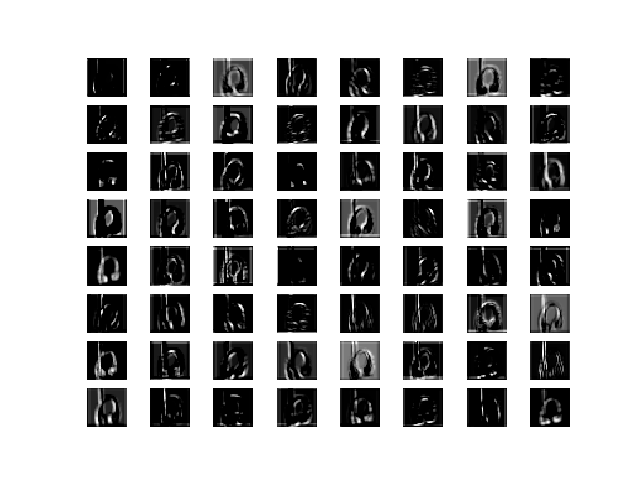}
		\caption{\scriptsize First layer, CNN using FBR}
	\end{subfigure}
	\begin{subfigure}[b]{0.48\linewidth}
		\includegraphics[width=\linewidth]{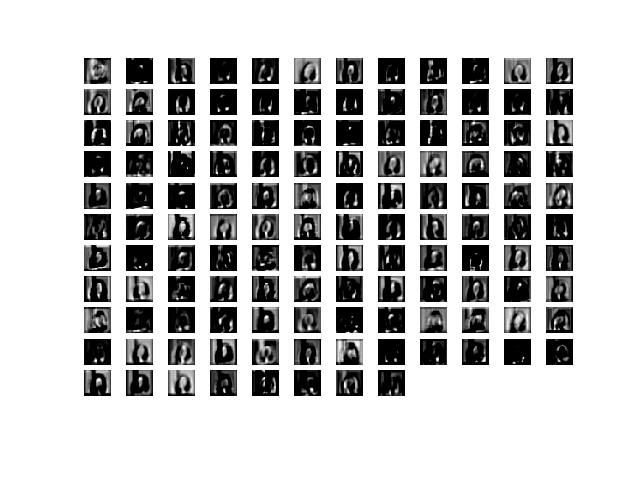}
		\caption{\scriptsize Third layer, baseline model}
	\end{subfigure}
	\begin{subfigure}[b]{0.48\linewidth}
		\includegraphics[width=\linewidth]{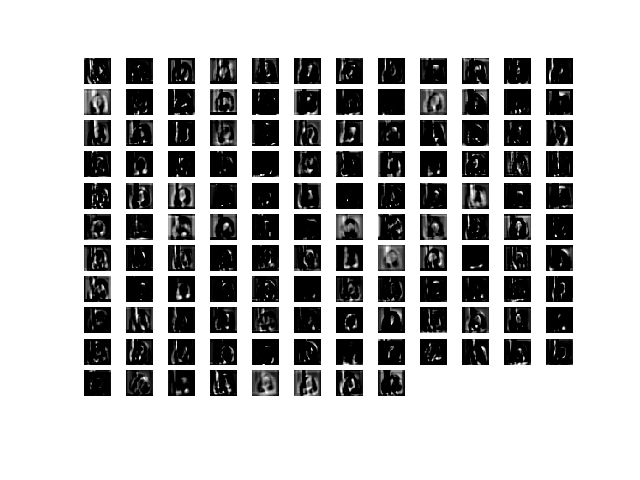}
		\caption{\scriptsize Third layer, CNN using FBR}
	\end{subfigure}
	\caption{Feature maps for the second test image}
	\label{fig:testsample2_featuremaps}
	\end{figure}

	The feature maps of layer 1 and layer 3 for the baseline and the FBR method are displayed in Figure \ref{fig:testsample1_featuremaps} and Figure \ref{fig:testsample2_featuremaps} respectively. The similar feature maps of the baseline model are marked in    
	these figures

	As can be easily observed in the figures, the feature maps of the FBR method are more sparse than those of the baseline model without regularization.  This increased sparsity improves the robustness of the FBR method.  Also, we bring the reader's attention to interpreting the feature maps of the FBR method in Figures \ref{fig:testsample1_featuremaps} and \ref{fig:testsample2_featuremaps}.  Thanks to the Gabor filters included in the regularization filter bank, the FBR method extracts features of strong directionality and high frequency that may explain the superior classification performance of the FBR method.

	\begin{figure}
		\centering
		\begin{subfigure}[b]{0.45\linewidth}
			\includegraphics[width=\linewidth]{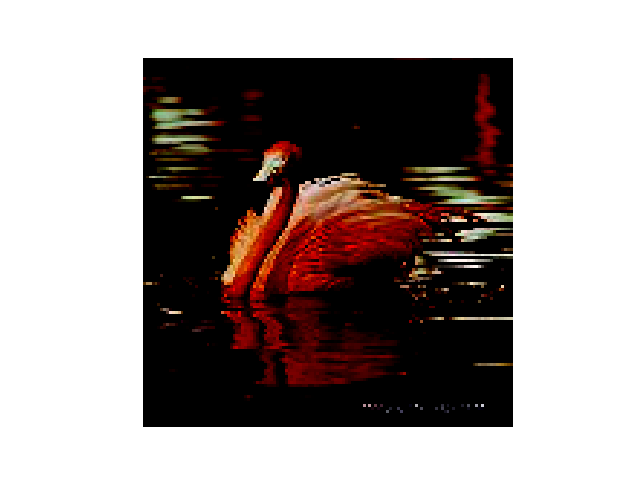}
			\caption{\scriptsize FBR prediction: "flamingo", $\ell_2$ Prediction: "ibis"}
		\end{subfigure}
		\begin{subfigure}[b]{0.45\linewidth}
			\includegraphics[width=\linewidth]{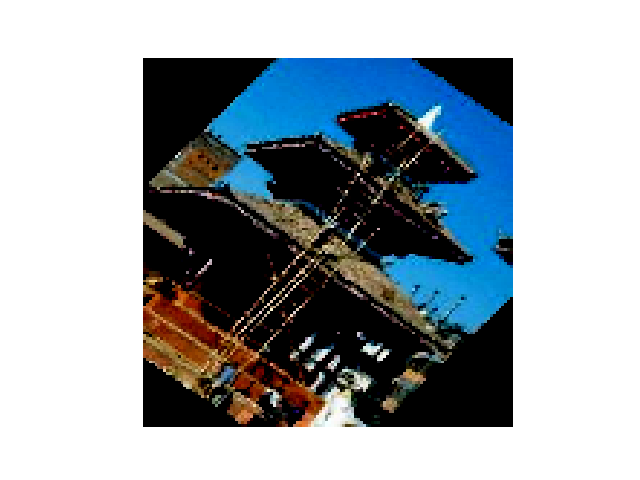}
			\caption{\scriptsize FBR prediction: "pagoda", $\ell_2$ Prediction: "accordion"}
		\end{subfigure}
		\caption{Example images}
		\label{fig:testsample34}
	\end{figure}

\begin{figure}
		\centering
		\begin{subfigure}[b]{0.48\linewidth}
			\includegraphics[width=\linewidth]{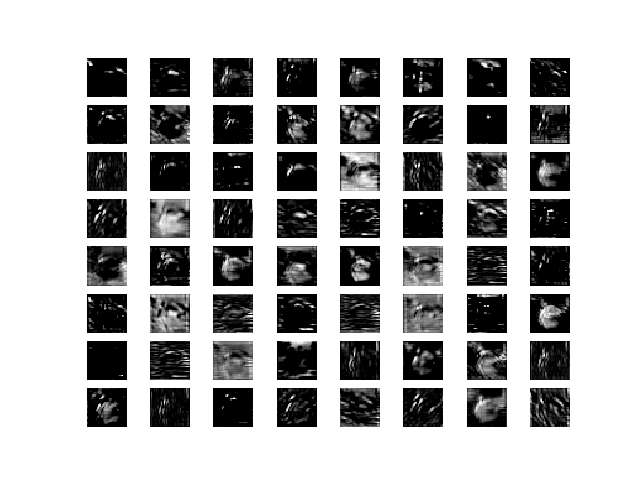}
			\caption{\scriptsize First layer, baseline model with $\ell_2$}
		\end{subfigure}
		\begin{subfigure}[b]{0.48\linewidth}
			\includegraphics[width=\linewidth]{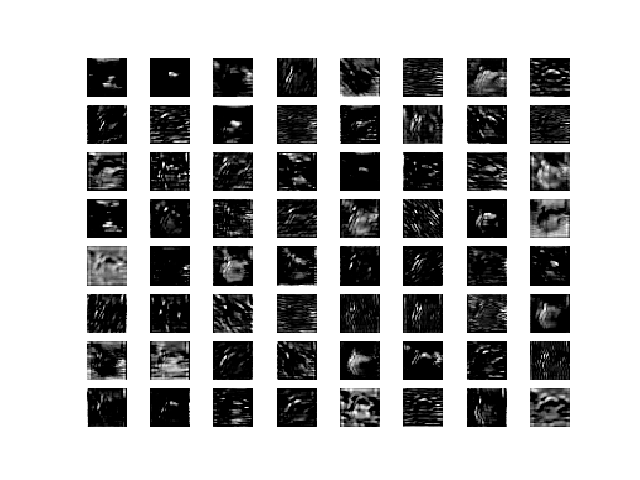}
			\caption{\scriptsize First layer, CNN using FBR}
		\end{subfigure}
		\begin{subfigure}[b]{0.48\linewidth}
			\includegraphics[width=\linewidth]{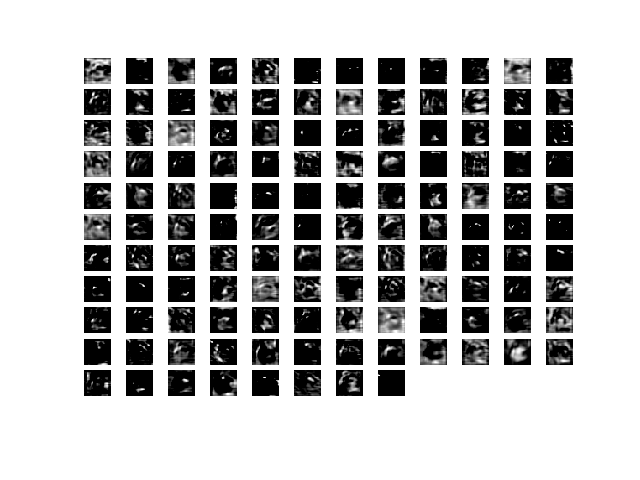}
			\caption{\scriptsize Third layer, baseline model with $\ell_2$}
		\end{subfigure}
		\begin{subfigure}[b]{0.48\linewidth}
			\includegraphics[width=\linewidth]{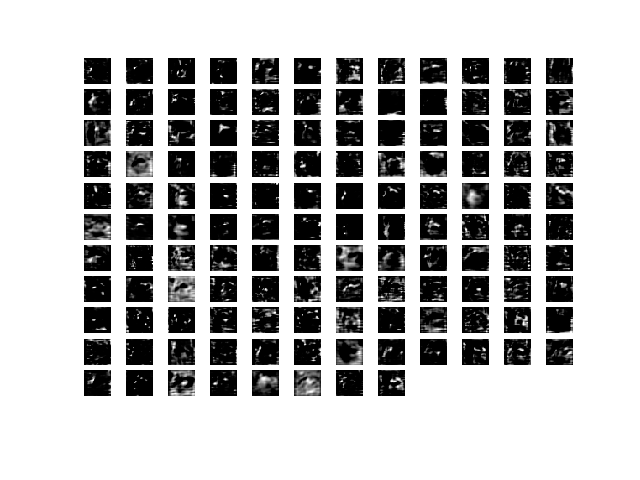}
			\caption{\scriptsize Third layer, CNN using FBR}
		\end{subfigure}
		\caption{Feature maps for the first test sample (FBR in comparison with $\ell_2$)}
		\label{fig:testsample3_featuremaps}
	\end{figure}

\begin{figure}
	\centering
	\begin{subfigure}[b]{0.48\linewidth}
		\includegraphics[width=\linewidth]{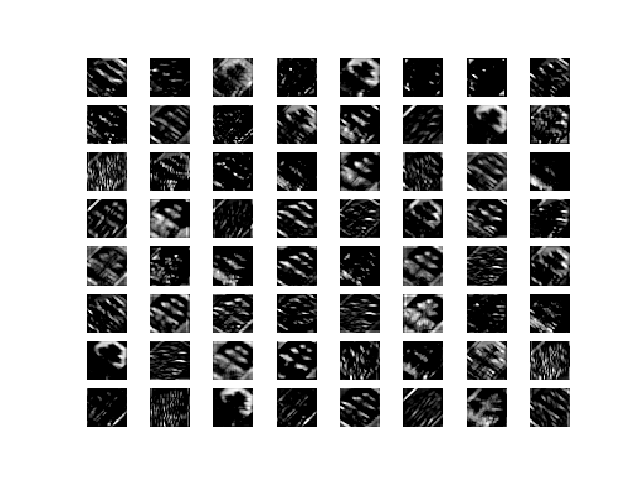}
		\caption{\scriptsize First layer, baseline model with $\ell_2$}
	\end{subfigure}
	\begin{subfigure}[b]{0.48\linewidth}
		\includegraphics[width=\linewidth]{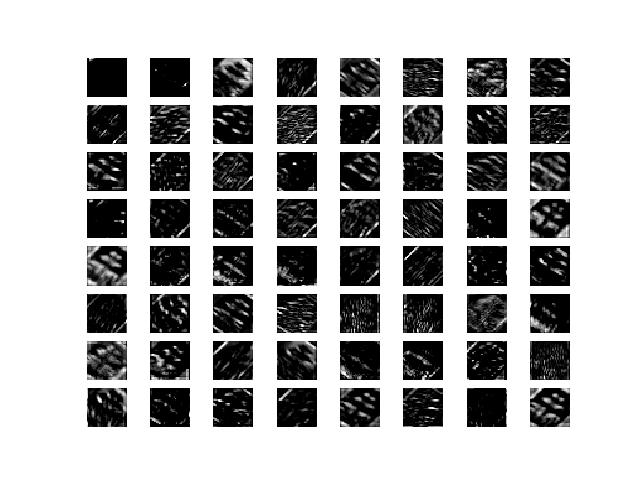}
		\caption{\scriptsize First layer, CNN using FBR}
	\end{subfigure}
	\begin{subfigure}[b]{0.48\linewidth}
		\includegraphics[width=\linewidth]{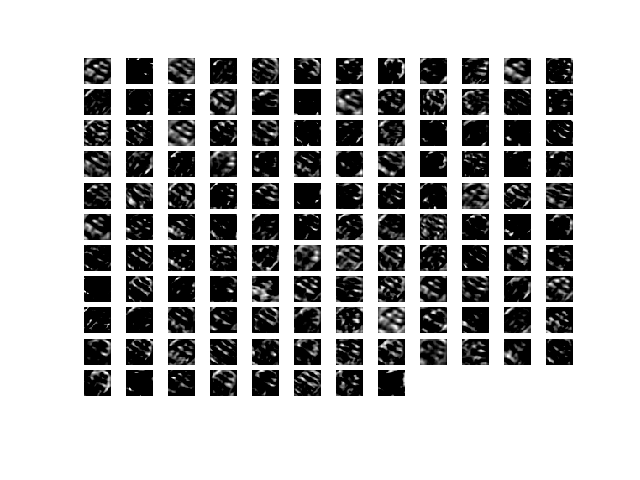}
		\caption{\scriptsize Third layer, baseline model with $\ell_2$}
	\end{subfigure}
	\begin{subfigure}[b]{0.48\linewidth}
		\includegraphics[width=\linewidth]{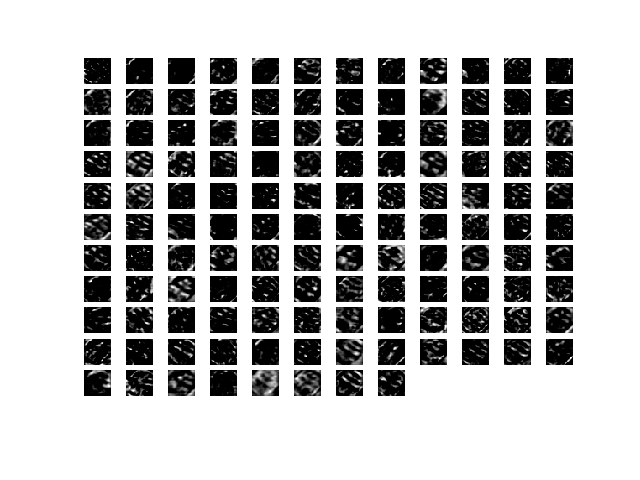}
		\caption{\scriptsize Third layer, CNN using FBR}
	\end{subfigure}
	\caption{Feature maps for the first test sample (FBR in comparison with $\ell_2$)}
	\label{fig:testsample4_featuremaps}
	\end{figure}

	Next, we compare the FBR method and the $\ell_2$ regularization.  Again, two sample images of the CalTech 101 dataset are selected and shown in Figures \ref{fig:testsample34}.
	For these two images the FBR method correctly classifies, whereas the $\ell_2$ regularization does not.
	%The test samples are displayed in Figures \ref{fig:testsample3} and \ref{fig:testsample4}. 
	The feature maps of layer 1 and layer 3 for the two methods are shown in Figures \ref{fig:testsample3_featuremaps} and \ref{fig:testsample4_featuremaps}.

	Here, the observations are very similar to what we discussed about Figures \ref{fig:testsample1_featuremaps} and \ref{fig:testsample2_featuremaps}.
	The feature maps of the FBR method appear to be sparser and exhibit greater discriminating power in high frequency and directionality than the $\ell_2$ regularization.  This explains the superior performance of the former over the latter.

	\section{Conclusion}

	Regularization techniques are widely used to prevent DCNNs from overfitting.  While the importance of regularization is generally accepted, no previously existing explicit regularization techniques take into account the spatial correlation of the weights of a convolution kernel in DCNNs.  This oversight has been addressed and it is corrected by our novel approach of filter bank regularization of DCNNs.  This regularization approach allows us to incorporate into the network training process interpretable feature extractors such as Gabor filters to improve the convergence, robustness and generality of DCNNs.
	
	\clearpage
	
	{\small
		\bibliographystyle{ieee}
		\bibliography{fbr}
	}
	
\end{document}